\documentclass{easychair}

\usepackage{doc}
\usepackage{booktabs}
\usepackage{float} 
\usepackage{amsmath, amssymb}
\usepackage{subcaption}

\title{Detecting Localized Deepfakes: How Well Do Synthetic Image Detectors Handle Inpainting?}

\author{
Serafino Pandolfini\thanks{Corresponding author}
\and
Lorenzo Pellegrini
\and
Matteo Ferrara
\and
Davide Maltoni
}

\institute{
University of Bologna, Italy\\
\email{serafino.pandolfini2@unibo.it}, 
\email{l.pellegrini@unibo.it}, 
\email{matteo.ferrara@unibo.it}, 
\email{davide.maltoni@unibo.it}
}

\authorrunning{S. Pandolfini, L. Pellegrini, M. Ferrara, and D. Maltoni}

\authorrunning{Mokhov, Sutcliffe and Voronkov}

\titlerunning{Detecting Localized Deepfakes}

\begin{document}

\maketitle

\begin{abstract}
The rapid progress of generative AI has enabled highly realistic image manipulations, including inpainting and region-level editing. These approaches preserve most of the original visual context and are increasingly exploited in cybersecurity-relevant threat scenarios. While numerous detectors have been proposed for identifying fully synthetic images, their ability to generalize to localized manipulations remains insufficiently characterized. This work presents a systematic evaluation of state-of-the-art detectors, originally trained for the deepfake detection on fully synthetic images, when applied to a distinct challenge: localized inpainting detection. The study leverages multiple datasets spanning diverse generators, mask sizes, and inpainting techniques. Our experiments show that models trained on a large set of generators exhibit partial transferability to inpainting-based edits and can reliably detect medium- and large-area manipulations or regeneration-style inpainting, outperforming many existing ad hoc detection approaches.
\end{abstract}


\setcounter{tocdepth}{2}

%
%

\section{Introduction}

The rapid advancement of generative AI has fundamentally reshaped the digital media landscape, enabling the creation of highly realistic synthetic imagery at an unprecedented scale. While these technologies offer powerful creative and professional capabilities, they have also amplified the spread of manipulated content intended to mislead, deceive, or imitate reality. This threat is further amplified by the proliferation of image inpainting models, which can seamlessly remove, modify, or insert visual elements into authentic images with minimal perceptual artifacts. Originally developed for image restoration, inpainting now facilitates subtle yet highly effective manipulations that pose significant risks in cybersecurity-critical domains, including coordinated disinformation campaigns, political influence operations, social engineering, and digital fraud.
Beyond classical inpainting, recent advances in image editing and image-to-image generative pipelines, such as prompt-based editing, object replacement, background regeneration, and region-guided diffusion, enable far more sophisticated manipulations that merge characteristics of both inpainting and full image generation. These systems can locally regenerate large image regions, rewrite semantic content, or even produce fully edited images while preserving the global structure and altering key elements in ways that remain almost imperceptible to human observers. From a forensic perspective, such manipulations represent an evolution of inpainting: they still rely on regenerating selected regions while maintaining coherence with the original image, but they operate at much larger scales and introduce transformations that blur the boundary between “edited” and “synthetic.” This emerging spectrum of localized-to-global modifications defines a new frontier in media manipulation, expanding the capabilities available to malicious actors and significantly complicating traditional detection strategies.
In recent years, synthetic content has been increasingly exploited to fabricate evidence, distort public narratives, and undermine trust in authentic digital media. Unlike fully synthetic images, inpainted or selectively regenerated images preserve most of the original visual context, making them particularly difficult to detect and highly effective for deceptive purposes. This emerging class of “localized deepfakes” expands the attack surface for adversaries and raises urgent concerns for platforms, institutions, and investigators that rely on trustworthy visual information.

In response, the research community has introduced a wide range of detection strategies, from handcrafted forensic analyses and frequency-domain artifact detection to patch-based classifiers and vision models fine-tuned to distinguish authentic from manipulated content. However, as with deepfake and synthetic image detection more broadly, reported performance often varies significantly depending on factors such as the choice of training generators, data preprocessing, augmentation strategies, and architectural biases. This lack of standardization complicates method comparison and obscures which design principles truly enhance generalization, particularly for inpainting detection, where manipulated regions may be small, context-dependent, and visually indistinguishable from their surroundings.

In this work, we present a comprehensive study of AI-generated image detection, with a particular focus on image inpainting, image editing, and localized manipulations, framed within the context of cybersecurity and media integrity. Following the principles established in AI-GenBench~\cite{pellegrini2025aigenbench}, we design an evaluation protocol tailored to inpainting and localized-editing forensics. Our protocol spans a variety of generators, mask dimensions, and manipulation types, providing a controlled environment for assessing the robustness of current detection approaches. This setting allows us to explore how well detectors trained for conventional real–fake classification on fully synthetic images transfer to more subtle, localized manipulations such as inpainting.

Rather than introducing a new detection model, our study evaluates state-of-the-art detectors originally trained on AI-GenBench~\cite{pellegrini2025aigenbench} for detecting fully synthetic images and analyzes their robustness when applied to localized inpainted content. We investigate the factors that most strongly influence generalization performance, including the size of the inpainted region, the underlying manipulation algorithm, and the characteristics of different generative models. By comparing detection behavior across these dimensions and comparing it to previous results on fully synthetic images, we identify patterns that reveal how inpainting-based manipulations differ from fully synthetic content in the eyes of current detectors. Through this targeted analysis, our work provides actionable insights for designing future forensic systems, highlighting failure modes and opportunities to improve robustness against localized attacks.

This paper is organized as follows. Section~\ref{sec:related_work} introduces the background and relevant related work. Section~\ref{sec:experimental_setup} details the experimental setup, including the inpainting datasets used, the training procedure, and the input pipeline. Section~\ref{sec:results} presents the experimental results from multiple dimensions of analysis. Section~\ref{sec:conclusions} summarizes our findings and highlights open research questions and directions for future work.

\section{Related Work}
\label{sec:related_work}
Detecting AI-assisted image manipulations increasingly extends beyond traditional deepfakes to include inpainting, localized edits, and partial image regeneration. These techniques preserve most of the original visual context while subtly altering specific regions, making them significantly harder to detect than fully synthetic images. In this section, we review the state-of-the-art in two closely related areas: (i) detection methods, focusing on architectural choices, forensic cues, and algorithmic strategies used to identify manipulations, and (ii) benchmarks and datasets, which define evaluation protocols, manipulation types, and generator diversity for assessing the robustness and generalization of detection systems.

\subsection{Inpainting Techniques}
Modern inpainting and localized editing methods can be broadly grouped into two methodological categories based on their usage in the generation process: (i) masked inpainting, or(ii) full regeneration.

\paragraph{Masked (pure) inpainting}-
One of the earliest approaches to inpainting relies on a binary mask to constrain the region to be modified. The objective is to reconstruct or hallucinate plausible content within the masked area while preserving global visual coherence. This category includes:
\begin{itemize}
    \item classical inpainting algorithms, such as Telea's method~\cite{telea2004animage},
    \item feed-forward convolutional architectures, such as LaMa~\cite{suvorov2022resolution} and MAT~\cite{li2022mat},
    \item diffusion-based inpainting models, such as Stable Diffusion 2~\cite{Rombach_2022_CVPR} and BrushNet~\cite{ju2024brushnet}.
\end{itemize}

\paragraph{Full regeneration}-
Fully regenerative methods are becoming increasingly prevalent, as they are often the default inpainting approach in simplified AI-systems, such as chat-based interfaces. These methods replace large masked regions or entire semantic components of an image using powerful generative priors. In doing so, they may substantially alter scene semantics, lighting conditions, or texture statistics while maintaining a high degree of perceptual realism. Examples of fully regenerative methods include:
\begin{itemize}
    \item diffusion-based region regeneration models, such as Stable Diffusion 2 (SD2)~\cite{Rombach_2022_CVPR} and Stable Diffusion XL (SDXL)~\cite{Rombach_2022_CVPR},
    \item text-guided editing pipelines, including Adobe Firefly~\cite{AdobeFirefly2025} and PowerPaint~\cite{zhuang2024powerpaint},
    \item large generative models, such as FLUX.1 schnell, FLUX.1 dev, and FLUX.1 filldev~\cite{BlackForestFlux2024}.
\end{itemize}
Image regeneration often introduces unwanted modifications, including identity alterations, changes in lighting conditions, the introduction of noticeable graininess, and a tendency toward cartoonization, among others.

These two categories pose different levels of difficulty for forensic analysis. Masked inpainting typically introduces subtle inconsistencies confined to the manipulated region, often resulting in anomalous noise patterns relative to the rest of the image. In contrast, semantic regeneration may produce globally coherent outputs that exhibit distributional shifts, challenging detectors that rely on local artifacts while potentially introducing image-wide signatures.

\subsection{Detection Methods}
Early forensic techniques primarily relied on handcrafted cues and statistical inconsistencies. Representative examples include Noiseprint~\cite{cozzolino2018noiseprint}, which models camera-specific noise residuals to localize anomalous regions, and frequency-domain or wavelet-based analyses that have been recently adapted to detect artifacts introduced by diffusion-based inpainting.

More recent approaches leverage deep learning for segmentation or classification. CTNet targets local tampering through patch-wise transformer reasoning, while DiffSeg~\cite{frick2024diffseg} introduces multi-feature segmentation specifically designed for diffusion-based inpainting detection. FakeShield~\cite{xu2025fakeshield} further advances the field by integrating multi-modal forensic cues and robust post-processing strategies to improve localized manipulation detection.

\subsection{Benchmarks and Datasets}
Benchmarks for inpainting and localized manipulation forensics remain considerably less mature than those developed for full-image deepfake or synthetic image detection. Existing datasets vary in scale, manipulation diversity, mask design, and realism, and often cover only a limited subset of inpainting algorithms. 

The Inpainting Localization Evaluation Dataset~\cite{wu2021inaintinglocalizationdataset} provides pixel-level annotations for manipulations generated using a combination of classical and deep-learning-based inpainting algorithms. The dataset includes results from foundational deep inpainting methods such as Contextual Attention~\cite{CA}, Gated Convolution~\cite{GC}, ShiftNet~\cite{SH}, EdgeConnect~\cite{EC}, and Region Normalization~\cite{RN}, alongside classical algorithms including Telea’s fast marching method~\cite{TE}, Navier–Stokes inpainting~\cite{NS}, and several patch-based or structure-guided approaches~\cite{PM,SG,LR}. While these methods were representative of the state of the art between 2017–2020, they predate modern diffusion and transformer-based inpainting models and therefore exhibit less realistic regeneration capabilities compared to contemporary systems. Although smaller in scale compared to more recent datasets, it remains a widely used benchmark for assessing segmentation-based approaches for inpainting localization.

DEFACTO~\cite{gael2019defacto} is a large-scale dataset designed for generic image manipulation detection, including splicing, copy-move, and inpainting. Its inpainting subset provides a variety of object-removal scenarios; however, most manipulations rely on classical or non-generative restoration techniques. Consequently, DEFACTO is less suitable for evaluating robustness against contemporary generative inpainting pipelines. 

DeepfakeArt~\cite{aboutalebi2024deepfakeart} is primarily oriented toward manipulation detection in artistic and stylized imagery. Although it includes localized alterations and generative transformations, its non-photographic domain limits its applicability to inpainting forensics on natural-image.

InpaintCOCO~\cite{roesch2024inpaintcoco} contains inpainted versions of COCO images generated using neural inpainting models, along with ground-truth masks for the removed regions. The dataset primarily emphasizes object-removal scenarios and masked inpainting pipelines, but includes fewer large-scale semantic regenerations compared to more recent benchmarks.

Beyond the Brush~\cite{bertazzini2024beyondthebush} targets modern generative editing workflows, including region-level editing, semantic modifications, and object-guided content generation. Although not exclusively focused on inpainting, portions of the dataset involve localized content replacement and regeneration, making it highly relevant to inpainting detection research.

TGIF~\cite{mareen2024tgif} and TGIF-2~\cite{TGIF2_2024} provide standardized resources for evaluating tampering localization and object-removal detection under modern generative editing pipelines. Unlike earlier inpainting benchmarks, both datasets incorporate manipulations produced by state of the art diffusion-based and text-guided inpainting systems, including models such as Adobe Firefly~\cite{AdobeFirefly2025} and various Flux1 models~\cite{BlackForestFlux2024}. The datasets further includes cases where the editing model regenerates the entire image rather than only the masked region, an emerging challenge for localized forensics, as traditional inpainting localization methods often fail under full-image regeneration.

BR-Gen~\cite{cai2025zooming} is a recent large-scale dataset designed for localized generative forgeries, emphasizing both object-level manipulations and broader scene edits such as sky, ground, or background regions. Although the dataset relies on slightly older but still widely used generative models—including LaMa~\cite{suvorov2022resolution}, BrushNet~\cite{ju2024brushnet}, PowerPaint~\cite{zhuang2024powerpaint}, SDXL~\cite{Rombach_2022_CVPR}, and MAT~\cite{li2022mat}, it offers a highly diverse set of masks and scene-aware annotations produced through an automated perception–creation–evaluation pipeline. This process ensures semantic coherence and high visual realism across the dataset.

\section{Experimental setup}
\label{sec:experimental_setup}
Our objective is to establish a baseline for detecting AI-generated inpainting and localized manipulations using models originally trained for binary real-versus-fake classification on fully synthetic images. Specifically, we assess whether state-of-the-art deepfake detectors trained on AI-GenBench can generalize to the inpainting and regeneration scenarios. 

We adopt a standard image-level detection formulation. Given an input image $I \in \mathbb{R}^{H \times W \times 3}$, the detector outputs a scalar probability $p_{\theta}(I)$ representing the likelihood that the image contains any AI-generated content. No spatial supervision is provided, and the detector is not informed about the location or size of manipulated regions. The task is therefore strictly binary: the model must decide whether the image contains any form of inpainting or regeneration.

To investigate architectural effects and to cover both convolutional and transformer-based paradigms, we evaluate two representative backbones previously considered in AI-GenBench: ResNet-50 CLIP~\cite{radford2021resnet50} by OpenAI and DINOv2~\cite{oquab2024dinov2} and DINOv3~\cite{siméoni2025dinov3}, both in their ViT-L variants.
\subsection{Training Procedure}
The two models are trained on AI-GenBench~\cite{pellegrini2025aigenbench} following the optimized designed choices outlined in~\cite{pellegrini2025generalized}, including optimization parameters, augmentation strategies, and model-specific hyperparameters. Key components include:

\begin{itemize}
    \item \emph{Resize-based preprocessing} - Input images are resized to the native resolution of the model without cropping.
    \item \emph{Augmentation pipeline} - Training uses the evaluation-based augmentation suite from~\cite{pellegrini2025generalized}, which includes multi-pass JPEG compression and limited perturbations such as Gaussian noise addition, grayscale conversion, rotations, and flipping.
    \item \emph{Supervision} - Only binary real-versus-fake supervision is used. No inpainting-specific data is included during training, nor any labels related to the model used to generate the inpainted region.
\end{itemize}

This protocol ensures that any performance on inpainting detection arises from generalization rather than from exposure to localized manipulations during training.

\subsection{Testing Datasets}
To construct our evaluation corpus, we aggregate real and manipulated images from established inpainting and localized-editing datasets, ensuring broad coverage across manipulation types, mask configurations, and generative pipelines.

\begin{figure}[!htbp]
    \centering
    \begin{subfigure}[t]{0.19\textwidth}
        \centering
        \includegraphics[width=\textwidth]{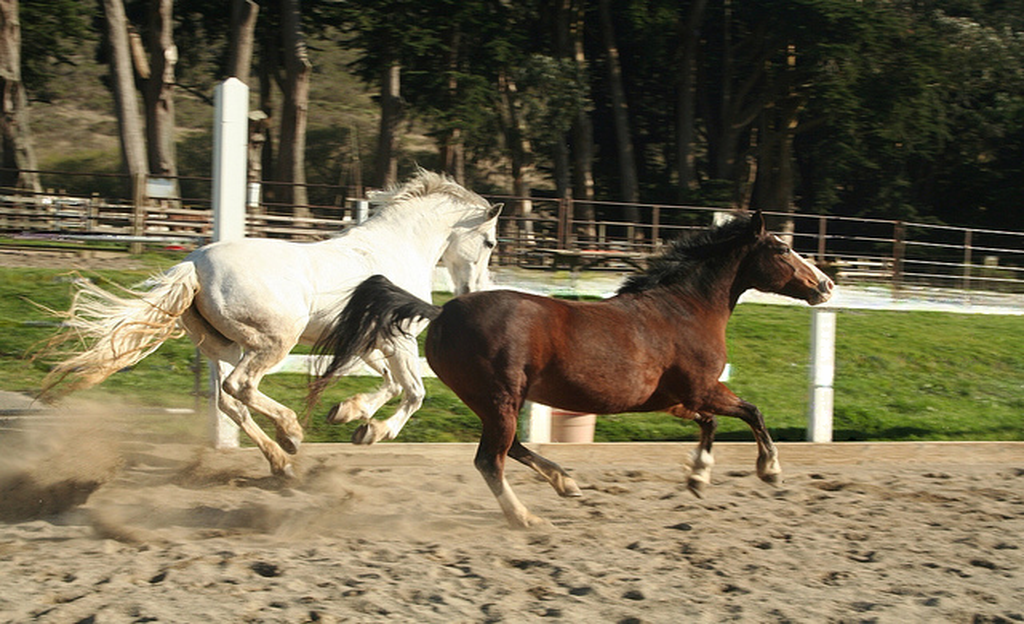}
        \caption{LaMa}
        \label{fig:lama-img}
    \end{subfigure}
    \hfill
    \begin{subfigure}[t]{0.19\textwidth}
        \centering
        \includegraphics[width=\textwidth]{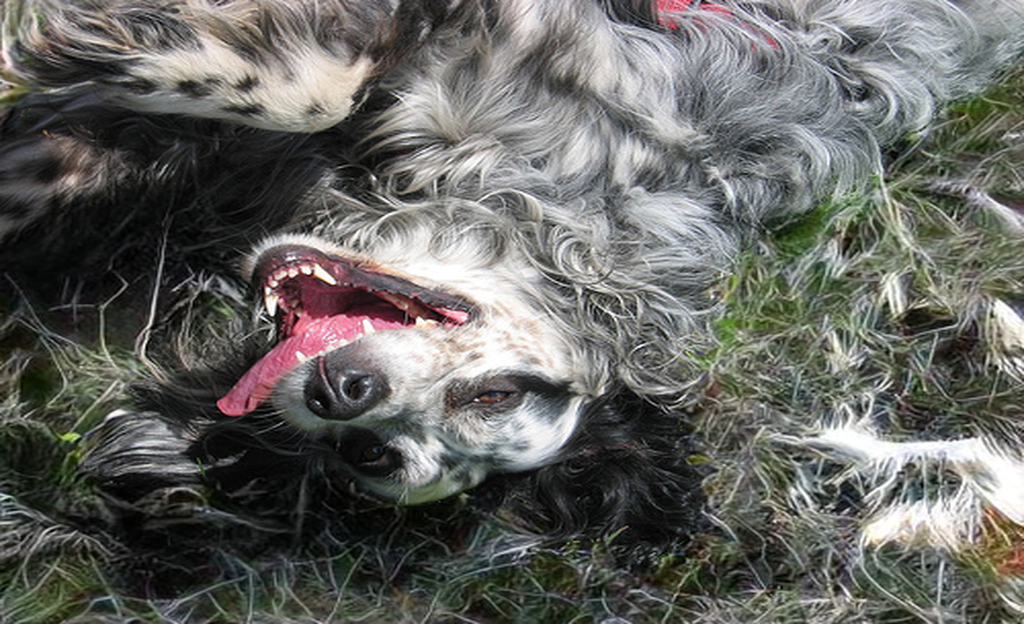}
        \caption{MAT}
        \label{fig:mat-img}
    \end{subfigure}
    \hfill
    \begin{subfigure}[t]{0.19\textwidth}
        \centering
        \includegraphics[width=\textwidth]{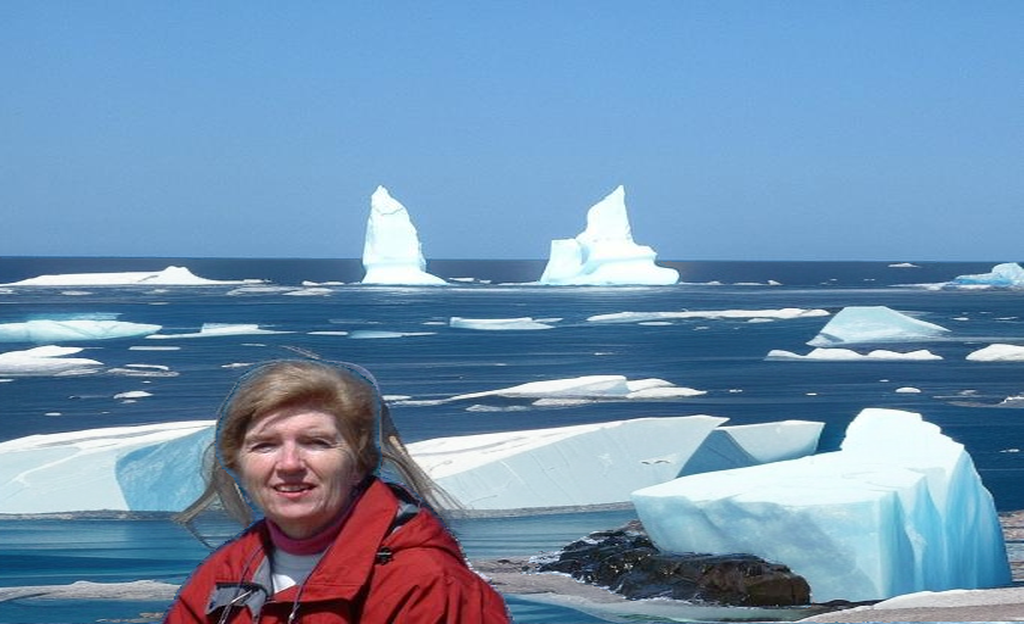}
        \caption{PowerPaint}
        \label{fig:powerpaint-img}
    \end{subfigure}
    \hfill
    \begin{subfigure}[t]{0.19\textwidth}
        \centering
        \includegraphics[width=\textwidth]{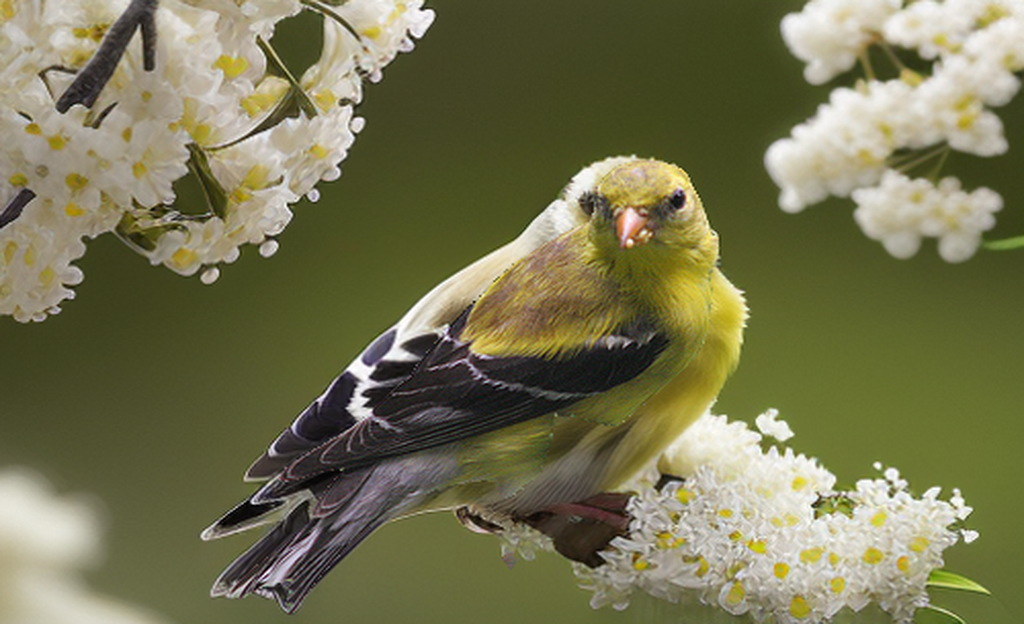}
        \caption{BrushNet}
        \label{fig:brushnet-img}
    \end{subfigure}
    \hfill
    \begin{subfigure}[t]{0.19\textwidth}
        \centering
        \includegraphics[width=\textwidth]{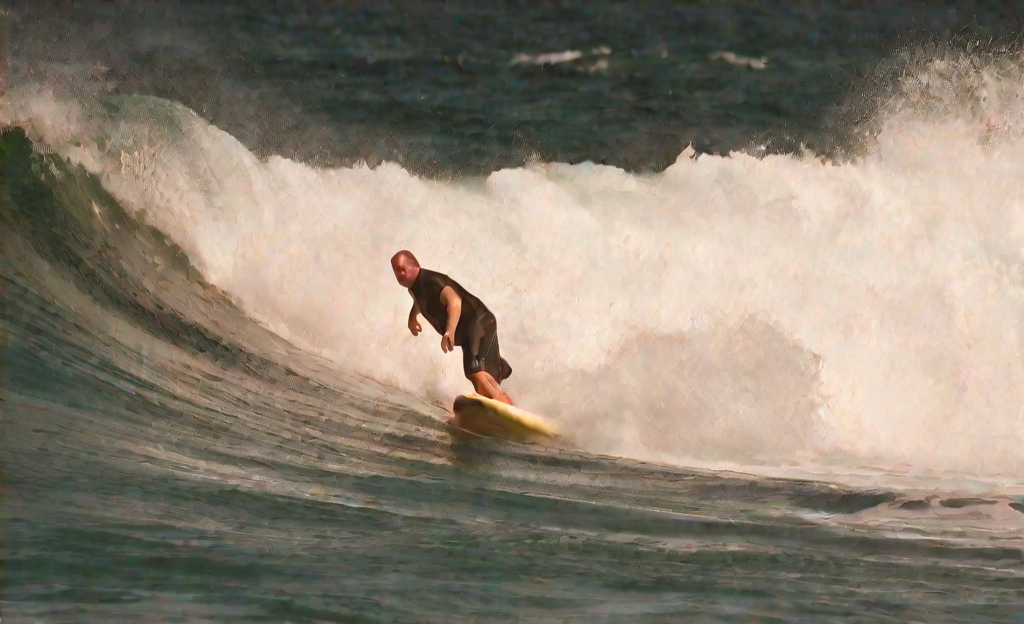}
        \caption{SD2}
        \label{fig:sdxl-img}
    \end{subfigure}

    \vspace{0.3cm}

    \begin{subfigure}[t]{0.19\textwidth}
        \centering
        \includegraphics[width=\textwidth]{images/SD2-example.png}
        \caption{SDXL}
        \label{fig:sd2-img}
    \end{subfigure}
    \hfill
    \begin{subfigure}[t]{0.19\textwidth}
        \centering
        \includegraphics[width=\textwidth]{images/firefly-example.png}
        \caption{Firefly}
        \label{fig:firefly-img}
    \end{subfigure}
    \hfill
    \begin{subfigure}[t]{0.19\textwidth}
        \centering
        \includegraphics[width=\textwidth]{images/flux1dev-example.png}
        \caption{Flux1 dev}
        \label{fig:flux1dev-img}
    \end{subfigure}
    \hfill
    \begin{subfigure}[t]{0.19\textwidth}
        \centering
        \includegraphics[width=\textwidth]{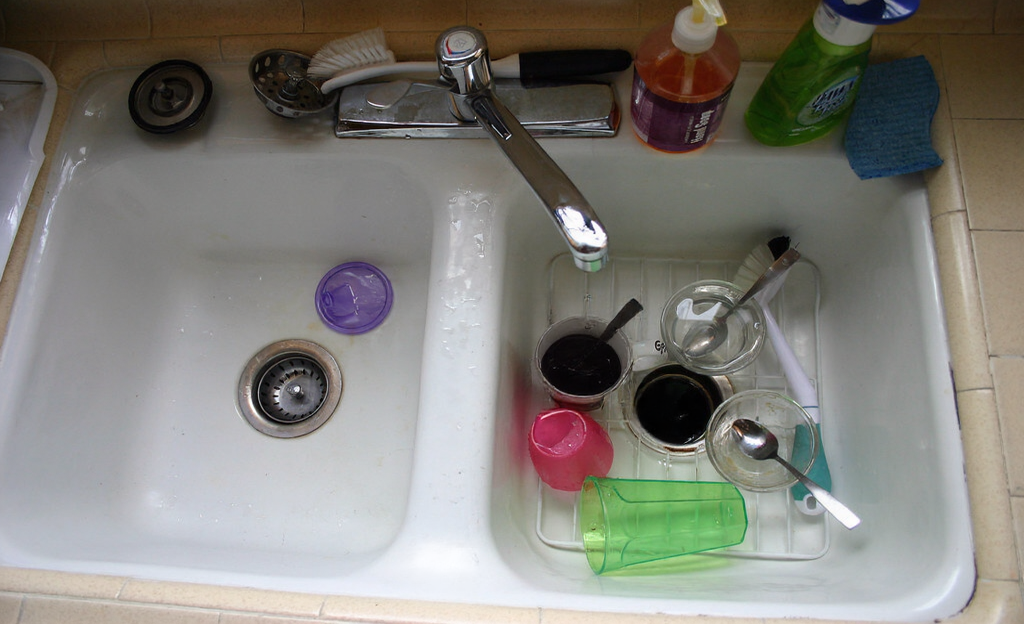}
        \caption{Flux1 filldev}
        \label{fig:flux1filldev-img}
    \end{subfigure}
    \hfill
    \begin{subfigure}[t]{0.19\textwidth}
        \centering
        \includegraphics[width=\textwidth]{images/flux1schnell-example.png}
        \caption{Flux1 schnell}
        \label{fig:flux1schnell-img}
    \end{subfigure}

    \caption{Inpainted images from different generators.}
    \label{fig:sample_images}
\end{figure}

Real images are taken from the unaltered subsets of BR-Gen~\cite{cai2025zooming}, which draws from COCO2017~\cite{lin2015microsoftcococommonobjects}, ImageNet~\cite{imagenet2009}, and Places~\cite{places2018}, as well as from TGIF~\cite{mareen2024tgif} and TGIF-2~\cite{TGIF2_2024}, both of which also originate from COCO2017. Manipulated samples are collected from BR-Gen, TGIF, and TGIF-2, capturing a heterogeneous range of modern generative inpainting techniques. These datasets collectively include forgeries produced by a diverse set of contemporary models. Figure~\ref{fig:sample_images} provides representative example images from the dataset.

\begin{figure}[H]
\centering
\includegraphics[width=0.95\linewidth]{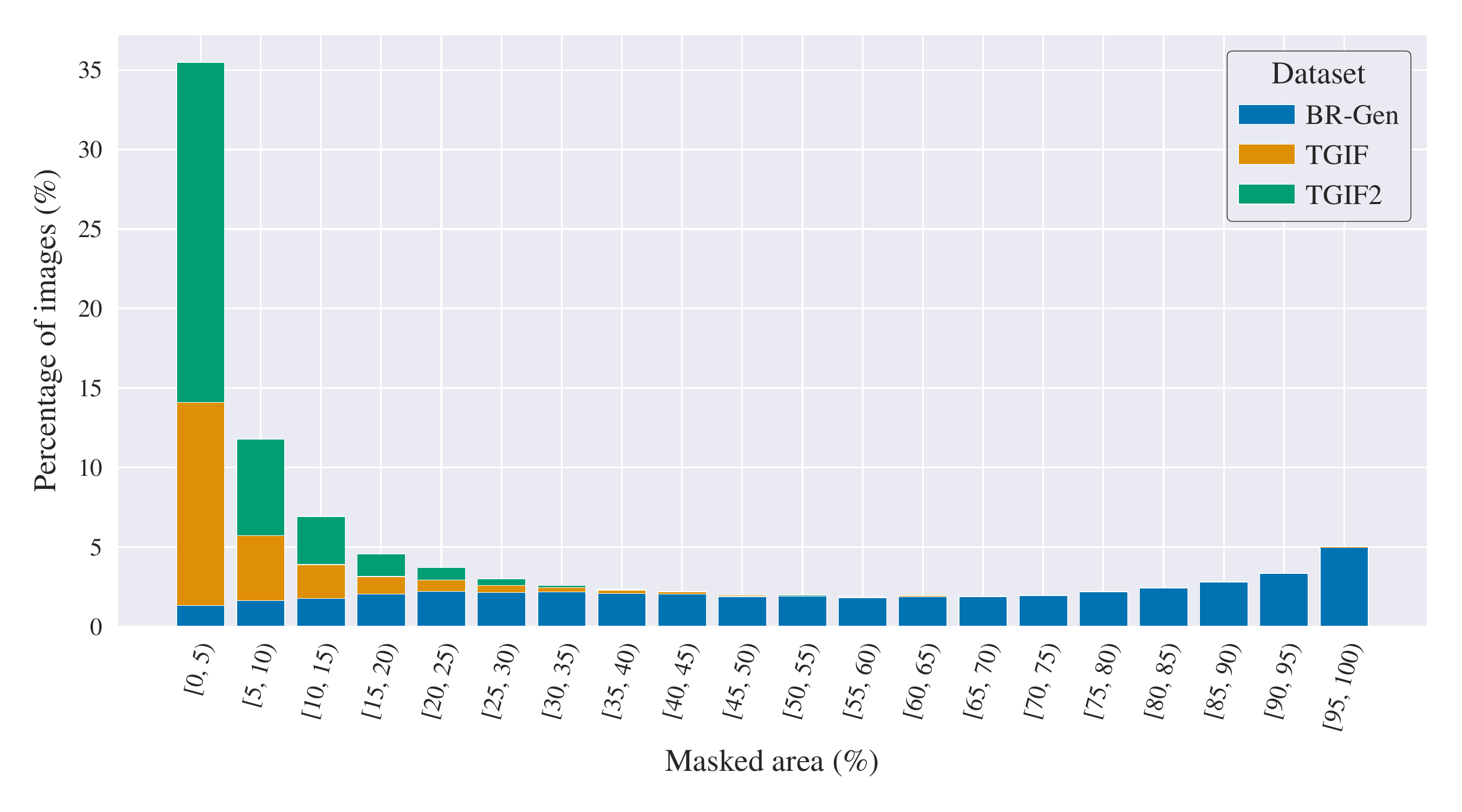}
\caption{Mask size distribution across the datasets.}
\label{fig:mask-size}
\end{figure}

\begin{figure}[H]
\centering
\includegraphics[width=1.0\linewidth]{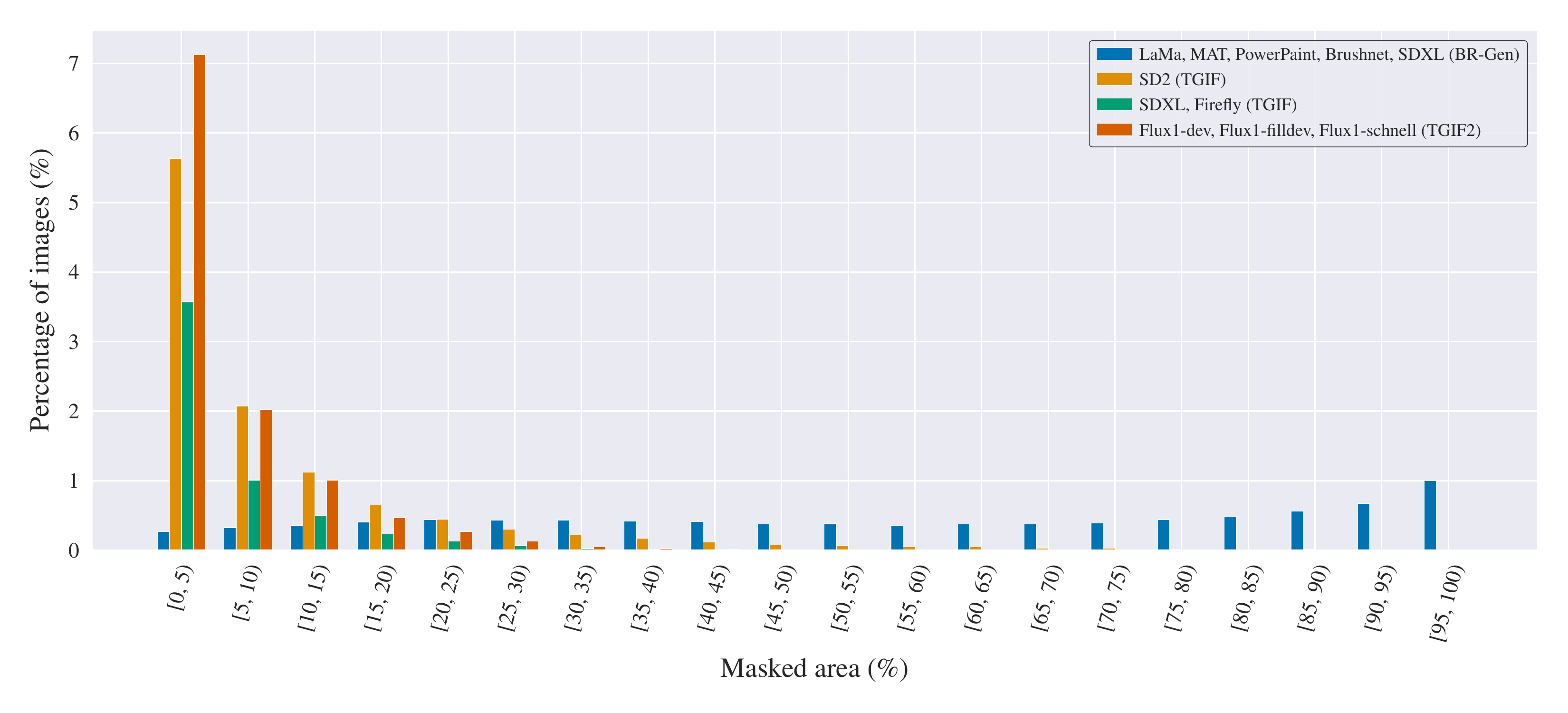}
\caption{Distribution of masked area percentages across representative generative models. For clarity and compactness, models with identical distributions in terms of number of images per bin have been grouped together, and only one representative model from each group is shown.}
\label{fig:mask-generators}
\end{figure}

The combined dataset inherits the rich diversity of mask shapes and scales present across its sources, spanning small occlusions, object-level removals, background and scene-region edits, and free-form semantic inpainting. As a result, the assembled dataset provides a broad and representative foundation for evaluating detection methods under realistic modern inpainting scenarios. An overview of the dataset composition is provided in Table~\ref{tab:generators}, which reports the number of inpainted and real images per source dataset. Figure~\ref{fig:mask-size} illustrates the mask size distributions for each dataset, while Figure~\ref{fig:mask-generators} shows the distribution of masked area percentages across generative models. Consistent with these distributions, Table~\ref{tab:mask_size_per_generator} summarizes the average mask size and standard deviation for each generator, revealing substantial variability across models. It is worth noting that the source datasets are unbalanced with respect to mask size (expressed as a percentage of of image area), and images manipulated with modern generative models tend to exhibit smaller masks.

\begin{table}[H]
\centering
\caption{Number of images in the datasets.}
\label{tab:generators}
\begin{tabular}{lcccc}
\toprule
Dataset & \# of inpainted images & \# of real images \\
\midrule
BR-Gen & 150000 & 15000 \\
TGIF & 74808 & 9372 \\
TGIF2 & 112464 & 6248 \\
\midrule
\textbf{Total} & \textbf{337272} & \textbf{30620} \\
\bottomrule
\end{tabular}
\end{table}

\begin{table}[H]
\centering
\small
\setlength{\tabcolsep}{6pt}
\caption{Average mask size and standard deviation for positive samples by manipulation generator.}
\label{tab:mask_size_per_generator}
\begin{tabular}{lcc}
\toprule
Generator & Avg Mask Size (\%) & Std (\%) \\
\midrule
LaMa & 56.4 & 30.1 \\
BrushNet & 56.4 & 30.1 \\
PowerPaint & 56.4 & 30.1 \\
MAT & 56.4 & 30.1 \\
SD2 & 9.8 & 12.9 \\
SDXL & 36.9 & 34.4 \\
Firefly & 5.5 & 6.4 \\
Flux.1 schnell & 5.5 & 6.4 \\
Flux.1 dev & 5.5 & 6.4 \\
Flux.1 filldev & 5.5 & 6.4 \\
\bottomrule
\end{tabular}
\end{table}

\subsection{Input pipeline}

The input pipeline closely follows the design validated in~\cite{pellegrini2025generalized}. Previous work has shown that classification performance and cross-generator generalization are maximized when detectors are trained and evaluated on entire images rather than on cropped or patch-based inputs. Accordingly:

\begin{itemize}
    \item \emph{Training} - Images are resized to the model's native input resolution and normalized according to the backbone specifications.
    \item \emph{Evaluation} - Each image is processed in its entirety using a single forward pass, with no test-time augmentation.
    \item \emph{Inference strategy} - No sliding-window or multi-crop inference is used, allowing us to measure the inherent sensitivity of the model to localized manipulations under realistic constraints.
\end{itemize}

\section{Results}
\label{sec:results}

For all experiments, we report the Area Under the Receiver Operating Characteristic curve (AUROC) as the primary evaluation metric. To analyze performance under different conditions, results are disaggregated by:
\begin{itemize}
    \item \textit{Testing dataset} - BR-Gen, TGIF, and TGIF2.
    \item \textit{Mask size} - expressed as a percentage of the image area.
    \item \textit{Inpainting type} - masked inpainting and semantic regeneration.
    \item \textit{Generator Identity} - SD2, SDXL, Firefly, Flux.1 dev, Flux.1 filldev, Flux.1 schnell, LaMa, BrushNet, PowerPaint, and MAT.
    \item \textit{Image compression} - uncompressed, JPEG Q80, and JPEG Q60.
\end{itemize}

\subsection{Evaluation on single datasets}
In this section, we compare the performance of our models (ResNet-50 CLIP, DINOv2 ViT-L/14 and DINOv3 ViT-L/16) against a range of baseline and state-of-the-art detection approaches reported in the literature across three datasets: BR-Gen, TGIF, and TGIF2. For each dataset.

\subsubsection{BR-Gen}

Table~\ref{tab:brgen-compare} reports AUROC results for detecting inpainted images within the BR-Gen dataset. Our ResNet-50 CLIP achieves performance comparable to other methods reported in the literature. In contrast, transformer-based self-supervised features substantially improve detection performance: DINOv2 ViT-L/14 significantly outperforms all prior baselines, achieving an AUROC above $0.83$, while DINOv3 ViT-L/16 slightly improves upon its predecessor, reaching an AUROC of $0.86$. These findings demonstrate the strong capability of self-supervised vision transformers in distinguishing inpainted content from authentic images.

\begin{table}[H]
\centering
\caption{AUROC scores for the BR-Gen dataset. Values for other methods are reported from~\cite{cai2025zooming}, while results marked as (ours) are obtained from our own evaluation under the same protocol. The highest and second-highest scores are highlighted in \textbf{bold} and \underline{underline}, respectively.}
\label{tab:brgen-compare}
\begin{tabular}{lc}
\toprule
Method & AUROC \\
\midrule
LGrad~\cite{tan2023learning}           & 0.64 \\
Dire~\cite{wang2023dire}           & 0.48 \\
FreqNet~\cite{tan2024freqnet}         & 0.47 \\
NPR~\cite{tan2023npr}             & 0.50 \\
FatFormer~\cite{liu2023fatformer}       & 0.60 \\
\midrule
\textbf{ResNet-50 CLIP (ours)}& 0.64 \\
\textbf{DINOv2 ViT-L/14 (ours)}   & \underline{0.83} \\
\textbf{DINOv3 ViT-L/16 (ours)}   & \textbf{0.86} \\
\bottomrule
\end{tabular}
\end{table}

\subsubsection{TGIF}

Following the evaluation protocol in~\cite{mareen2024tgif}, Table~\ref{tab:tgif-compare} presents AUROC results for detecting fully regenerated images within the TGIF dataset under different compression settings (uncompressed, JPEG Q80, and JPEG Q60). Our models maintain robust performance across all compression levels. Notably, DINOv2 ViT-L/14 and especially DINOv3 ViT-L/16 consistently achieve high AUROC values, demonstrating strong generalization to varying compression artifacts. The ResNet-50 CLIP model also exhibits competitive performance, particularly in moderately compressed images, indicating resilience to quality degradation.

\begin{table}[H]
\centering
\small
\setlength{\tabcolsep}{2pt}
\caption{AUROC scores for the TGIF dataset using only fully regenerated images. Results for baseline methods are taken from~\cite{mareen2024tgif}, while methods marked as (ours) are evaluated under the same protocol. The highest and second-highest scores in each column are highlighted in \textbf{bold} and \underline{underline}, respectively.}
\label{tab:tgif-compare}
\begin{tabular}{lcccccccc}
\toprule
& \multicolumn{2}{c}{Uncompressed} & \multicolumn{2}{c}{JPEG Q80} & \multicolumn{2}{c}{JPEG Q60} & \multicolumn{2}{c}{Average}\\
\cmidrule(lr){2-3} \cmidrule(lr){4-5} \cmidrule(lr){6-7} \cmidrule(lr){8-9}
Method & SD2 & SDXL & SD2 & SDXL & SD2 & SDXL &  SD2 & SDXL\\
\midrule
CNNDetect~\cite{wang2020cnndetect}   & 0.57 & 0.61 & 0.57 & 0.62 & 0.56 & 0.61 & 0.57 & 0.61 \\
DIMD~\cite{corvi2022dmid}        & \textbf{1.00} & \underline{0.94} & \textbf{1.00} & \textbf{0.95} & \textbf{0.99} & \textbf{0.92} & \textbf{1.00} & \textbf{0.94} \\
Dire~\cite{wang2023dire}        & 0.49 & 0.62 & 0.49 & 0.61 & 0.49 & 0.60 & 0.49 & 0.61\\
FreqDetect~\cite{frank2020freqnet}  & 0.50 & 0.40 & 0.72 & 0.71 & 0.71 & 0.72 & 0.64 & 0.61\\
UnivFD~\cite{ojha2024univfd}     & 0.82 & 0.80 & 0.73 & 0.78 & 0.70 & 0.73 & 0.75 & 0.77\\
Fusing~\cite{ju2022fusing}      & 0.55 & 0.47 & 0.52 & 0.59 & 0.53 & 0.61 & 0.53 & 0.56\\
GramNet~\cite{liu2020gramnet}     & 0.82 & 0.55 & 0.48 & 0.56 & 0.47 & 0.54 & 0.59 & 0.55\\
LGrad~\cite{tan2023learning}       & 0.85 & 0.83 & 0.51 & 0.57 & 0.51 & 0.64 & 0.62 & 0.68\\
NPR~\cite{tan2023npr}         & 0.51 & 0.67 & 0.50 & 0.30 & 0.51 & 0.68 & 0.51 & 0.55\\
RINE~\cite{koutlis2024rine}        & 0.89 & 0.89 & 0.79 & 0.88 & 0.75 & 0.85 & 0.81 & 0.87\\
DeFake~\cite{sha2023defake}      & 0.65 & 0.61 & 0.54 & 0.45 & 0.54 & 0.43 & 0.58 & 0.50\\
PatchCraft~\cite{zhong2024patchcraft}  & \underline{0.98} & \textbf{0.95} & 0.66 & 0.67 & 0.60 & 0.58 & 0.75 & 0.73\\
\midrule
\textbf{ResNet-50 CLIP (ours)} & 0.55 & 0.65 & 0.65 & 0.72 & 0.64 & 0.70 & 0.61 & 0.69\\
\textbf{DINOv2 ViT-L/14 (ours)}    & 0.91 & 0.86 & 0.90 & 0.85 & 0.88 & 0.83 & 0.90 & 0.85\\
\textbf{DINOv3 ViT-L/16 (ours)}    & 0.96 & 0.92 & \underline{0.95} & \underline{0.91} & \underline{0.93} & \underline{0.90} & \underline{0.95} & \underline{0.91}\\
\bottomrule
\end{tabular}
\end{table}

Table~\ref{tab:tgif-total} reports AUROC scores on the TGIF dataset when considering all manipulated images jointly, including both fully regenerated and spliced/inpainted samples. Overall performance is notably lower than in the fully regenerated setting (Table~\ref{tab:tgif-compare}), reflecting the increased difficulty of detecting localized manipulations, that often affect only a small portion of the image. As illustrated in Figure~\ref{fig:mask-size}, inpainted regions in TGIF frequently cover less than 20\% of the image area, resulting in weaker global cues.

Among the evaluated methods, DINOv3 ViT-L/16 achieves the highest AUROC, followed closely by DINOv2 ViT-L/14, indicating that large-scale self-supervised vision transformers retain a clear advantage even when multiple manipulation types are aggregated. In contrast, the ResNet-50 CLIP model performs closer to chance level, further highlighting the benefit of transformer-based representations for detecting subtle and heterogeneous image manipulations.

\begin{table}[H]
\centering
\caption{AUROC scores for the TGIF dataset.}
\label{tab:tgif-total}
\begin{tabular}{lc}
\toprule
Model & AUROC \\
\midrule
ResNet-50 CLIP & 0.54 \\
DINOv2 ViT-L/14 & 0.66\\
DINOv3 ViT-L/16 & \textbf{0.69}\\
\bottomrule
\end{tabular}
\end{table}

\subsubsection{TGIF2}

For TGIF2 (Table~\ref{tab:tgif2-total}), all evaluated models achieve AUROC scores close to random chance. This behavior can be partly attributed to the use of Flux1 generator variants, which produce images with significantly higher visual quality and fewer detectable artifacts compared to the generative models included in TGIF. As a result, the global image statistics captured by current feature representations provide very limited discriminative power for distinguishing real from synthetic content.

In addition, TGIF2 exhibits a manipulation mask area distribution similar to TGIF, where the majority of spliced or inpainted regions cover less than 20\% of the image area~\ref{fig:mask-size}. The prevalence of such localized and subtle modifications further increases detection difficulty, as global features are weakly affected by small manipulated regions. Compared to TGIF, where fully regenerated images can often be detected with high confidence, TGIF2 therefore poses a substantially harder detection problem due to the combination of stronger generative models and limited spatial extent of the manipulations.

Overall, these results highlight the limitations of existing feature-based detection approaches and underscore the need for representations that are both more sensitive to localized artifacts and more robust to advances in generative image quality.

\begin{table}[t]
\centering
\caption{AUROC scores for the TGIF2 dataset.}
\label{tab:tgif2-total}
\begin{tabular}{lc}
\toprule
Model & AUROC \\
\midrule
ResNet-50 CLIP & \textbf{0.53} \\
DINOv2 ViT-L/14 & 0.52\\
DINOv3 ViT-L/16 & 0.52\\
\bottomrule
\end{tabular}
\end{table}

\subsection{Effect of Mask Size}
\label{sec:eff_mask_size}

\begin{figure}[t]
\centering
\includegraphics[width=0.9\linewidth]{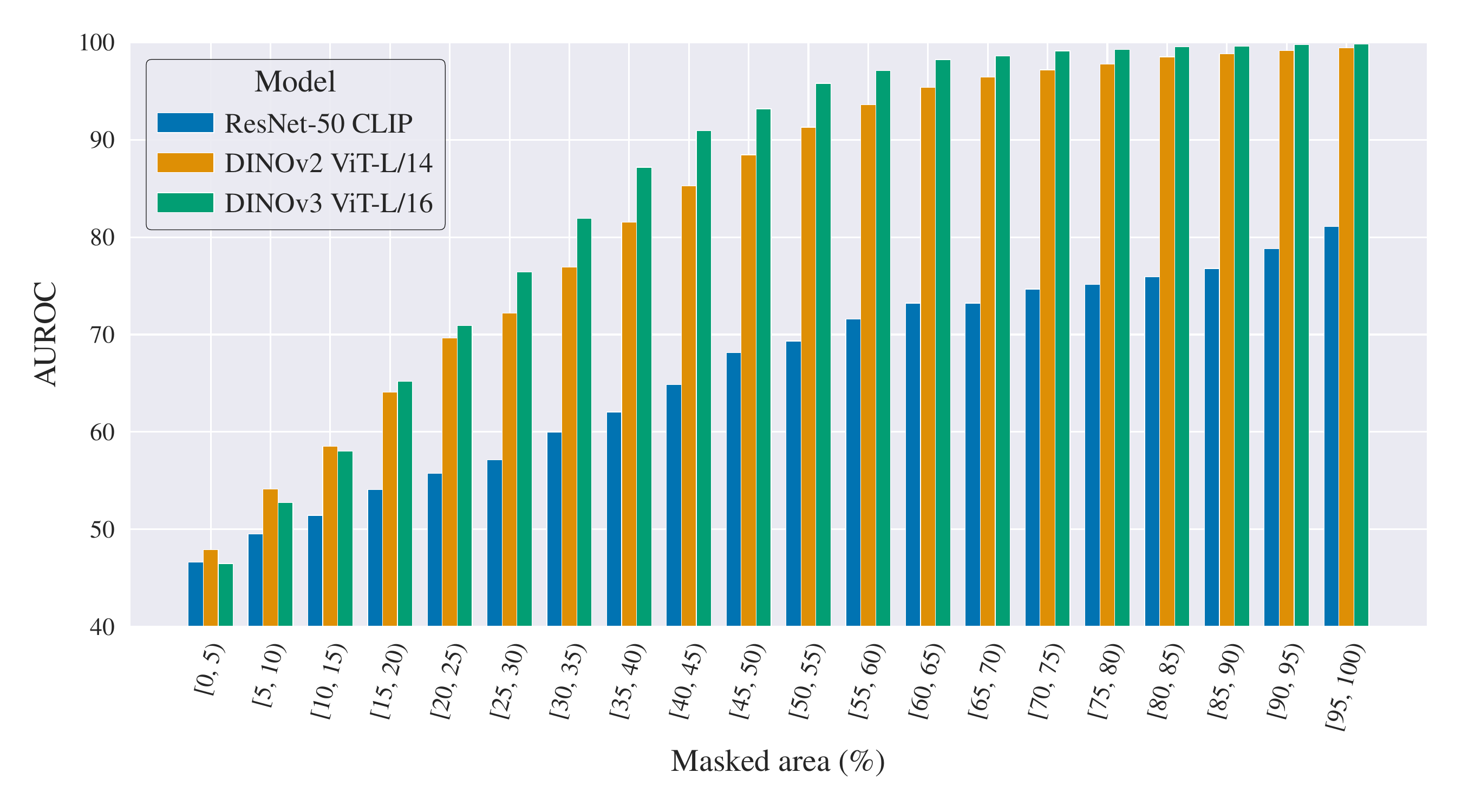}
\caption{AUROC as a function of mask size (percentage of image area modified). Larger regions yield more detectable artifacts.}
\label{fig:auroc-per-mask-size}
\end{figure}

To quantify the impact of manipulation scale, we evaluate detection performance across mask-size bins defined by the percentage of the image area that is replaced. Figure~\ref{fig:auroc-per-mask-size} reports the AUROC for each bin.

Detection accuracy correlates positively with mask size: for regions smaller than 5\% of the image, all detectors perform near chance. For regions greater than 20\% performance improves substantially, with DINOv3 consistently outperforming DINOv2 and ResNet-50, achieving higher AUROC scores, although both remain below their performance on fully generated images in AI-GenBench. For regions covering most of the image, DINOv3 reaches performance comparable to that observed on fully generated images at approximately $65\%$ mask size, whereas DINOv2 requires larger manipulated regions ($>80\%$) to achieve comparable similar results~\cite{pellegrini2025generalized}.

These findings suggest that current fully synthetic image detectors rely heavily on global artifacts, which are attenuated or absent in small localized edits.

\subsection{Effect of Inpainting Type}

We consider two categories of manipulations: masked (pure) inpainting and semantic full regeneration. 
Table~\ref{tab:inpainting-type} reports AUROC performance stratified by manipulation type.

\begin{table}[H]
\centering
\caption{Overall AUROC for detecting masked inpainting and full regeneration across all generators and mask sizes.}
\label{tab:inpainting-type}
\begin{tabular}{lcc}
\toprule
Model & Masked Inpainting & Full Regeneration \\
\midrule
ResNet-50 CLIP & 0.59 & 0.53 \\
DINOv2 ViT-L/14 & 0.70 & \textbf{0.60} \\
DINOv3 ViT-L/16 & \textbf{0.71} & \textbf{0.60} \\
\bottomrule
\end{tabular}
\end{table}

Results indicate that masked inpainting is generally more detectable, particularly for classical or feed-forward models that introduce localized texture inconsistencies. In contrast, full regeneration often yields globally coherent outputs that challenge detectors trained on full-image classification, sometimes producing near-chance performance for small masks (see Section~\ref{sec:eff_mask_size}).

\subsection{Effect of Generator Identity}

We further analyze cross-generator generalization by evaluating detectors separately on manipulations produced by different generative models. Table~\ref{tab:auroc_per_generator} reports AUROC performance stratified by generator.

\begin{table}[H]
\centering
\small
\setlength{\tabcolsep}{4pt}
\caption{AUROC by manipulation generator. Detection performance varies widely across generators, with diffusion-based models often proving the most challenging to identify.}
\label{tab:auroc_per_generator}
\begin{tabular}{lccc}
\toprule
Generator & ResNet-50 & DINOv2 & DINOv3 \\
\midrule
LaMa & 0.63 & 0.84 & \textbf{0.88} \\
BrushNet & 0.67 & 0.87 & \textbf{0.89} \\
PowerPaint & 0.69 & 0.88 & \textbf{0.90} \\
MAT & 0.70 & 0.86 & \textbf{0.89} \\
SD2 & 0.55 & \textbf{0.67} & \textbf{0.67} \\
SDXL & 0.64 & 0.79 & \textbf{0.83} \\
Firefly & 0.45 & \textbf{0.46} & 0.43 \\
Flux.1 schnell & \textbf{0.44} & 0.43 & 0.41 \\
Flux.1 dev & \textbf{0.46} & \textbf{0.46} & 0.43 \\
Flux.1 filldev & \textbf{0.49} & 0.48 & 0.45 \\
\bottomrule
\end{tabular}
\end{table}

A consistent observation is the difficulty of detecting manipulations produced by models of the Flux and Firefly families, particularly when masks are small. These models generate high-quality texture statistics that are distributionally closer to real content, thereby reducing detectable artifacts. Diffusion-based regenerators are often more difficult to distinguish than masked inpainting networks such as LaMa or MAT, which tend to leave structural or frequency-domain irregularities.

\subsection{Robustness to JPEG Compression}

Since a substantial fraction of online images undergo lossy recompression, we evaluate whether JPEG compression affects inpainting detectability. Following the compression settings used in prior work, we assess performance at multiple JPEG quality levels.

Results in Table~\ref{tab:compression} indicate that JPEG compression does not substantially affect detection accuracy. This finding is consistent with the training strategy proposed in~\cite{pellegrini2025generalized}, which incorporates aggressive multi-pass JPEG augmentation.

\begin{table}[H]
\centering
\caption{AUROC by JPEG compression.}
\label{tab:compression}
\begin{tabular}{lccc}
\toprule
Model & Uncompressed & JPEG Q80 & JPEG Q60\\
\midrule
ResNet-50 & 0.57 & 0.56 & 0.56 \\
DINOv2 & 0.67 & \textbf{0.67} & \textbf{0.67} \\
DINOv3 & \textbf{0.68} & \textbf{0.67} & 0.66 \\
\bottomrule
\end{tabular}
\end{table}

\subsection{Qualitative Analysis}

\begin{figure}[htbp]
    \centering

    \begin{minipage}[t]{0.48\textwidth}
        \centering
        \includegraphics[width=0.48\textwidth]{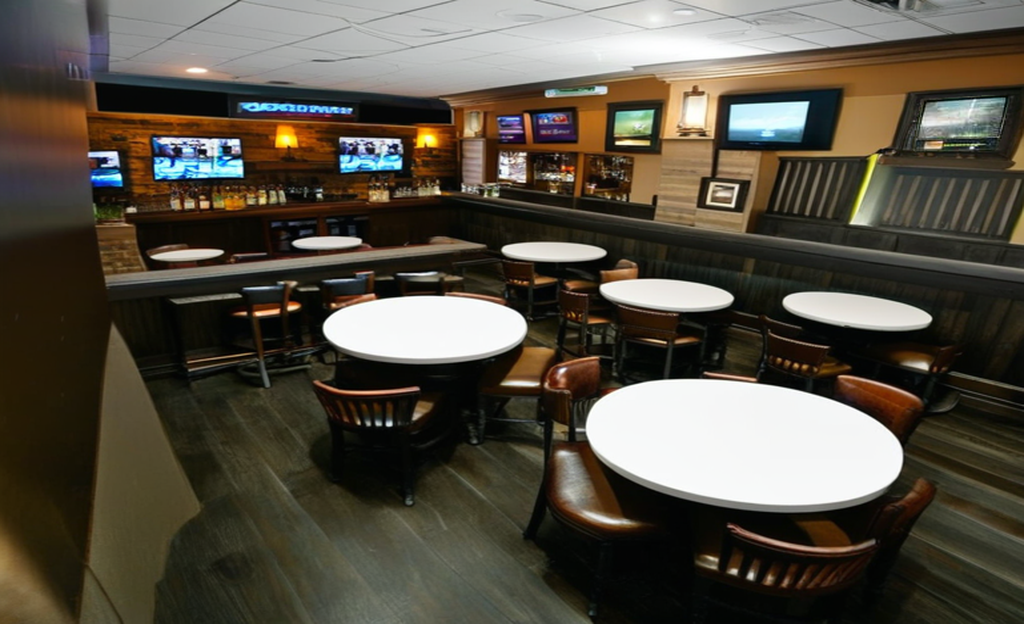}
        \includegraphics[width=0.48\textwidth]{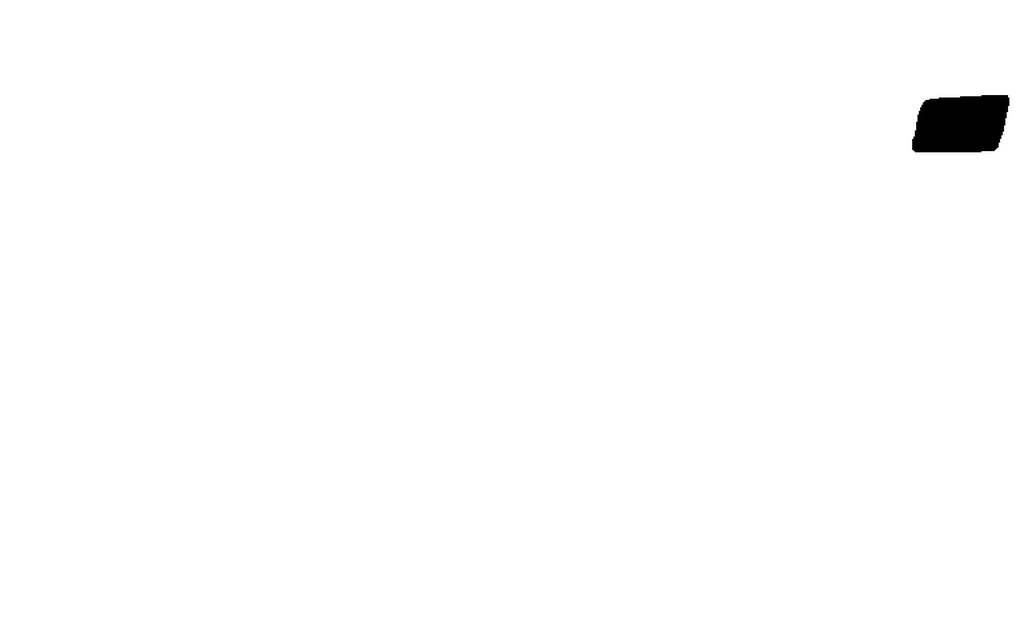}
        \label{fig:coppia1}
    \end{minipage}
    \hfill
    \begin{minipage}[t]{0.48\textwidth}
        \centering
       \includegraphics[width=0.48\textwidth]{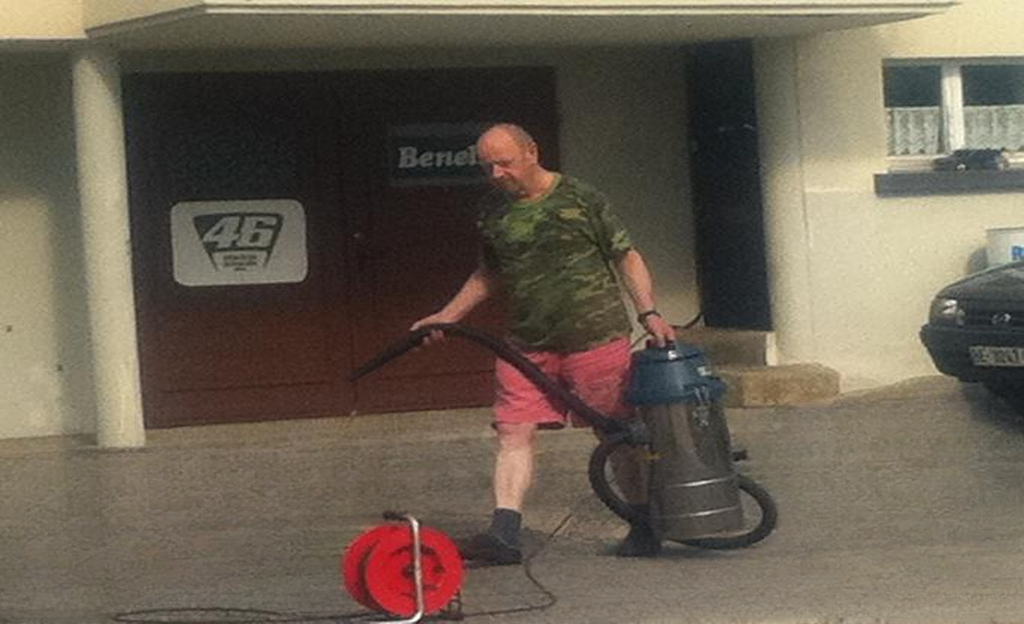}
        \includegraphics[width=0.48\textwidth]{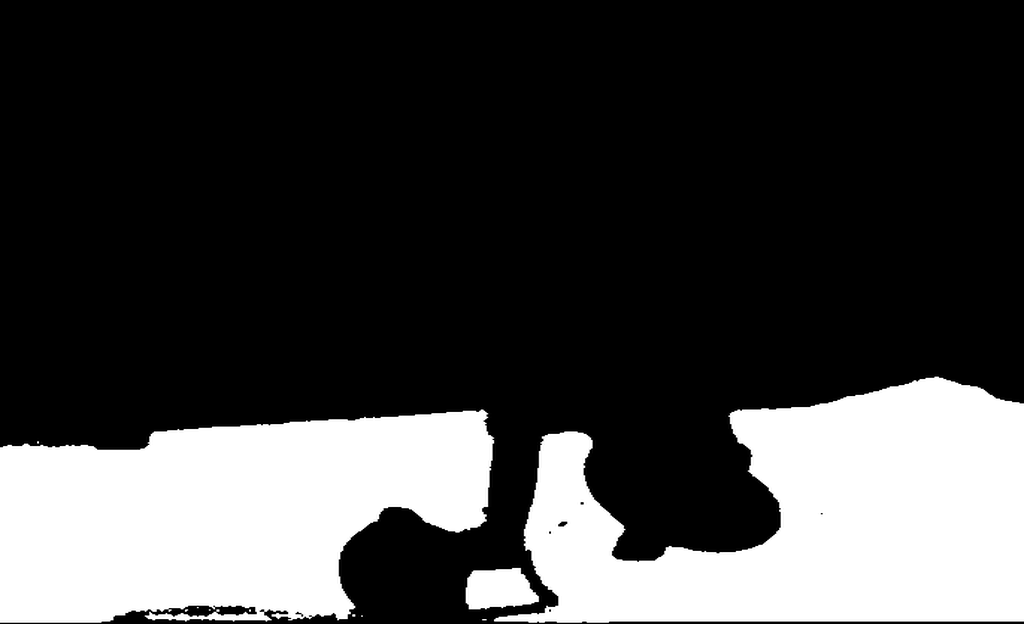}
        \label{fig:coppia2}
    \end{minipage}

    \vspace{0.3cm}

    \begin{minipage}[t]{0.48\textwidth}
        \centering
        \includegraphics[width=0.48\textwidth]{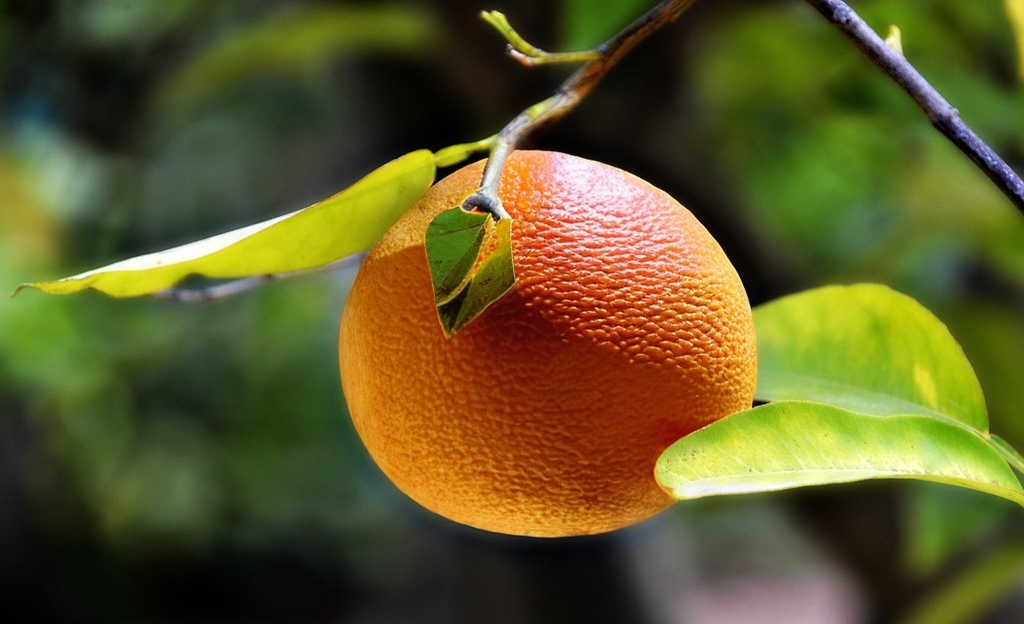}
        \includegraphics[width=0.48\textwidth]{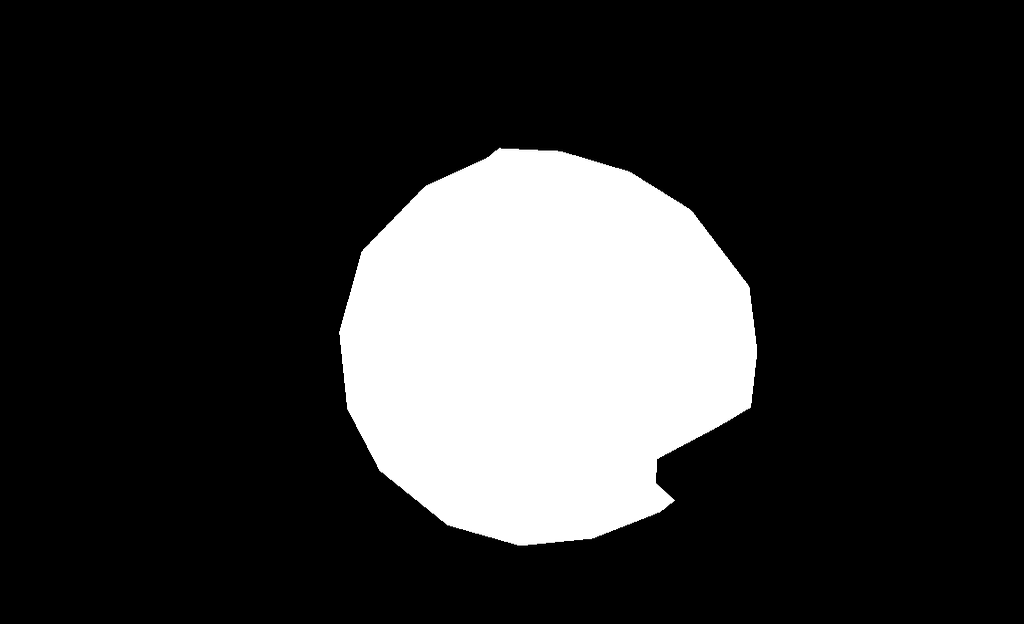}
        \label{fig:coppia3}
    \end{minipage}
    \hfill
    \begin{minipage}[t]{0.48\textwidth}
        \centering
         \includegraphics[width=0.48\textwidth]{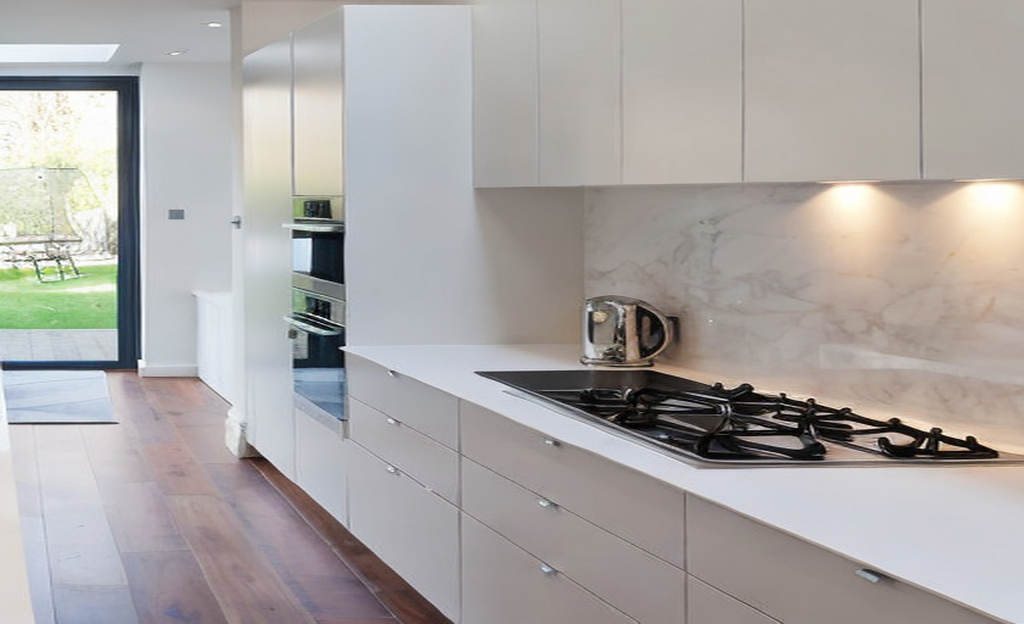}
        \includegraphics[width=0.48\textwidth]{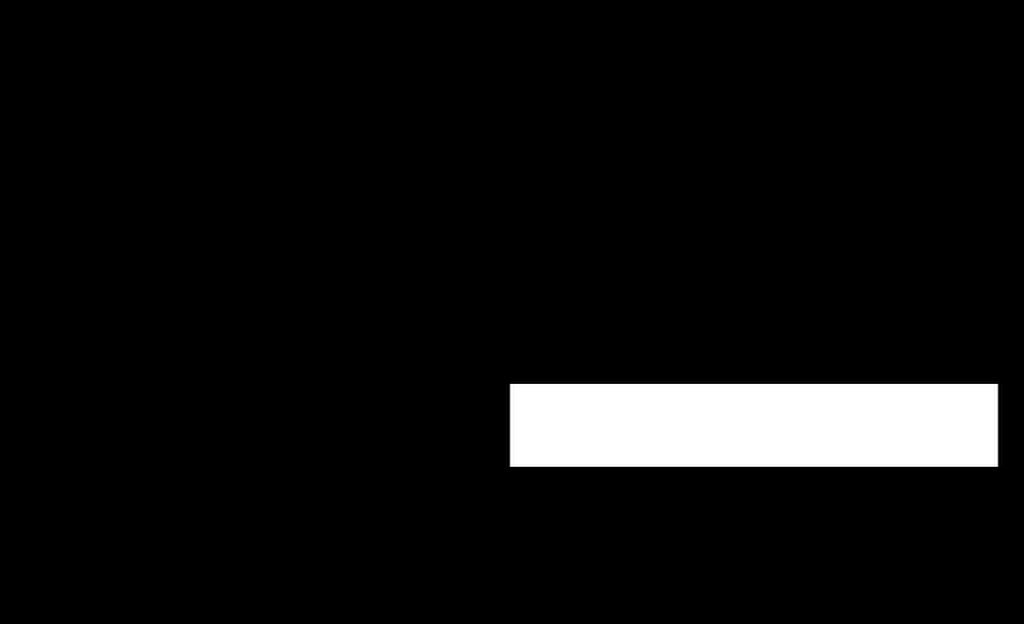}
        \label{fig:coppia4}
    \end{minipage}

    \vspace{0.3cm}

    \begin{minipage}[t]{0.48\textwidth}
        \centering
        \includegraphics[width=0.48\textwidth]{images/correct3_tgif2_fr.png}
        \includegraphics[width=0.48\textwidth]{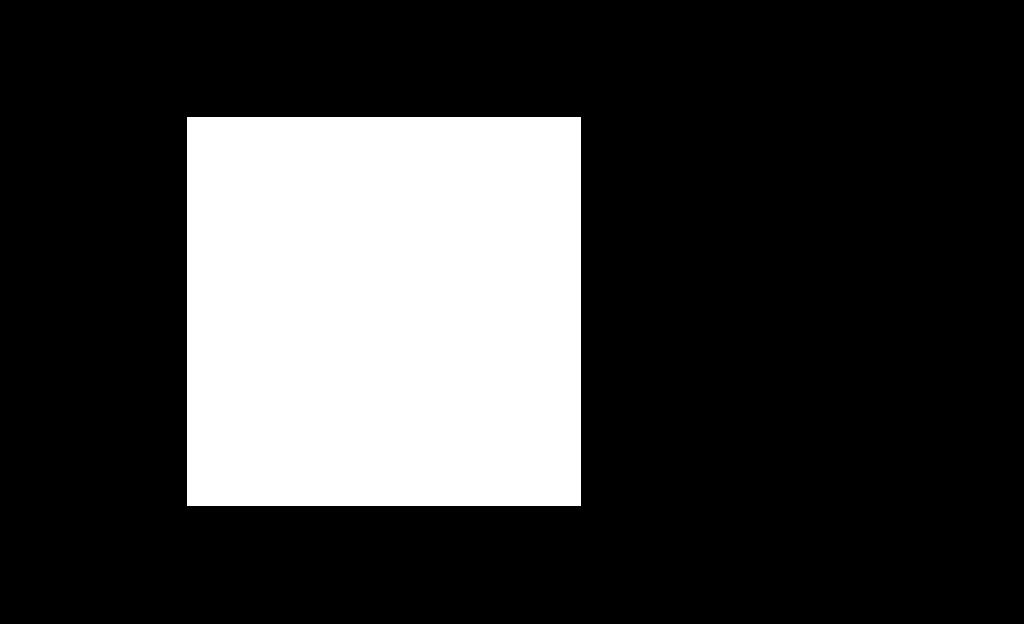}
        \label{fig:coppia5}
    \end{minipage}
    \hfill
    \begin{minipage}[t]{0.48\textwidth}
        \centering
        \includegraphics[width=0.48\textwidth]{images/fake3_tgif2_filldev_sp.png}
        \includegraphics[width=0.48\textwidth]{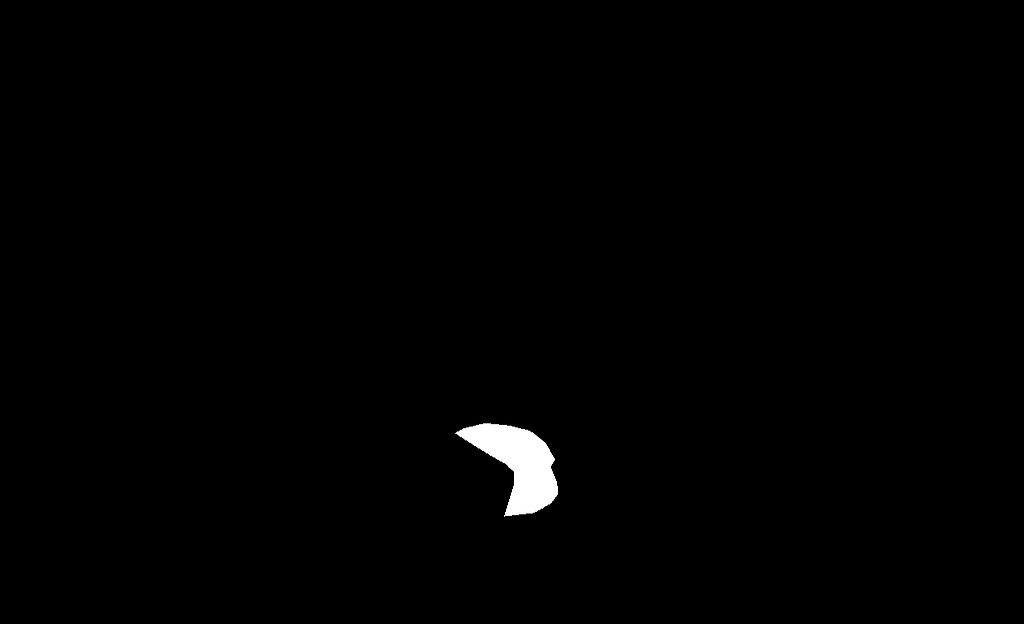}
        \label{fig:coppia6}
    \end{minipage}

    \caption{Qualitative examples of inpainted images with corresponding masks, illustrating successful detections (left) and failure cases (right) for the DINOv2 model.}
    \label{fig:qualitative}
\end{figure}

To better understand the limitations of current detectors, Figure~\ref{fig:qualitative} presents qualitative examples of images with low and high predicted manipulation scores.

Detectors tend to succeed on:  
\begin{itemize}
    \item manipulations that introduce clear boundary inconsistencies,  
    \item classical inpainting outputs with texture discontinuities,  
    \item large semantic changes not fully aligned with global context.
\end{itemize}

They struggle with:  
\begin{itemize}
    \item small inpainting masks,  
    \item high-quality diffusion-based regeneration (especially Flux models),  
    \item semantically coherent edits that preserve lighting and texture statistics.
\end{itemize}

\section{Conclusions}
\label{sec:conclusions}
This study examined the extent to which state-of-the-art pre-trained detectors for AI-generated content can generalize from fully synthetic images to localized inpainting. Our results show that detectors trained on a wide range of generators, such as those included in the AI-GenBench~\cite{pellegrini2025aigenbench}, exhibit competitive transferability to the inpainting detection task: even without explicit localization mechanisms (e.g., segmentation or bounding-box prediction), they can often identify whether an image contains inpainted regions.

Performance is particularly strong when the manipulated area is relatively large ($>50\%$) or when the inpainting approach involves full image regeneration. In these scenarios, our models can surpass state-of-the-art synthetic image detection mechanisms. However, we observe consistent performance degradation for fully regenerated images or when the edited region occupies only a small fraction of the image ($<20\%$), demonstrating that these cases remain challenging for models trained on the binary classification task (real vs fake).

Overall, the results indicate that, while broad generator exposure and robust augmentation during training enhance generalization, current classification-based detectors still struggle to detect small localized manipulations without specific detection or segmentation mechanisms. Future work should explore hybrid approaches that combine global classification with fine-grained localization to improve robustness against subtle inpainting attacks.

\section{Acknowledgments}

\begin{itemize}
\item We acknowledge, for Lorenzo Pellegrini, the support of the European funds from the Emilia-Romagna Region under the Fse+ 2021-2027 program.
\item We acknowledge ISCRA for awarding this project access to the LEONARDO supercomputer, owned by the EuroHPC Joint Undertaking, hosted by CINECA (Italy).
\end{itemize}

\bibliographystyle{plain}
\bibliography{easychair}

@InProceedings{mareen2024tgif,
  author="Mareen, Hannes and Karageorgiou, Dimitrios and Van Wallendael, Glenn and Lambert, Peter and Papadopoulos, Symeon",
  title="TGIF: Text-Guided Inpainting Forgery Dataset",
  booktitle="Proc. Int. Workshop on Information Forensics and Security (WIFS) 2024",
  year="2024"
}

@misc{cai2025zooming,
      title={Zooming In on Fakes: A Novel Dataset for Localized AI-Generated Image Detection with Forgery Amplification Approach}, 
      author={Lvpan Cai and Haowei Wang and Jiayi Ji and YanShu ZhouMen and Yiwei Ma and Xiaoshuai Sun and Liujuan Cao and Rongrong Ji},
      year={2025},
      eprint={2504.11922},
      archivePrefix={arXiv},
      primaryClass={cs.CV},
      url={https://arxiv.org/abs/2504.11922}, 
}

@INPROCEEDINGS{gael2019defacto,
  author={Mahfoudi, Gaël and Tajini, Badr and Retraint, Florent and Morain-Nicolier, Frédéric and Dugelay, Jean Luc and Pic, Marc},
  booktitle={2019 27th European Signal Processing Conference (EUSIPCO)}, 
  title={DEFACTO: Image and Face Manipulation Dataset}, 
  year={2019},
  volume={},
  number={},
  pages={1-5},
  doi={10.23919/EUSIPCO.2019.8903181}}

@InProceedings{suvorov2022resolution,
    author    = {Suvorov, Roman and Logacheva, Elizaveta and Mashikhin, Anton and Remizova, Anastasia and Ashukha, Arsenii and Silvestrov, Aleksei and Kong, Naejin and Goka, Harshith and Park, Kiwoong and Lempitsky, Victor},
    title     = {Resolution-Robust Large Mask Inpainting With Fourier Convolutions},
    booktitle = {Proceedings of the IEEE/CVF Winter Conference on Applications of Computer Vision (WACV)},
    month     = {January},
    year      = {2022},
    pages     = {2149-2159}
}

@article{telea2004animage,
author = {Alexandru Telea},
title = {An Image Inpainting Technique Based on the Fast Marching Method},
journal = {Journal of Graphics Tools},
volume = {9},
number = {1},
pages = {23--34},
year = {2004},
publisher = {Taylor \& Francis},
doi = {10.1080/10867651.2004.10487596},
}

@misc{pellegrini2025generalized,
      title={Generalized Design Choices for Deepfake Detectors}, 
      author={Lorenzo Pellegrini and Serafino Pandolfini and Davide Maltoni and Matteo Ferrara and Marco Prati and Marco Ramilli},
      year={2025},
      eprint={2511.21507},
      archivePrefix={arXiv},
      primaryClass={cs.CV},
      url={https://arxiv.org/abs/2511.21507}, 
}

@inproceedings{tan2023learning,
  title={Learning on Gradients: Generalized Artifacts Representation for GAN-Generated Images Detection},
  author={Tan, Chuangchuang and Zhao, Yao and Wei, Shikui and Gu, Guanghua and Wei, Yunchao},
  booktitle={Proceedings of the IEEE/CVF Conference on Computer Vision and Pattern Recognition},
  pages={12105--12114},
  year={2023}
}

@misc{wang2023dire,
      title={DIRE for Diffusion-Generated Image Detection}, 
      author={Zhendong Wang and Jianmin Bao and Wengang Zhou and Weilun Wang and Hezhen Hu and Hong Chen and Houqiang Li},
      year={2023},
      eprint={2303.09295},
      archivePrefix={arXiv},
      primaryClass={cs.CV},
      url={https://arxiv.org/abs/2303.09295}, 
}

@misc{tan2024freqnet,
      title={Frequency-Aware Deepfake Detection: Improving Generalizability through Frequency Space Learning}, 
      author={Chuangchuang Tan and Yao Zhao and Shikui Wei and Guanghua Gu and Ping Liu and Yunchao Wei},
      year={2024},
      eprint={2403.07240},
      archivePrefix={arXiv},
      primaryClass={cs.CV},
      url={https://arxiv.org/abs/2403.07240}, 
}

@misc{tan2023npr,
      title={Rethinking the Up-Sampling Operations in CNN-based Generative Network for Generalizable Deepfake Detection}, 
      author={Chuangchuang Tan and Huan Liu and Yao Zhao and Shikui Wei and Guanghua Gu and Ping Liu and Yunchao Wei},
      year={2023},
      eprint={2312.10461},
      archivePrefix={arXiv},
      primaryClass={cs.CV},
      url={https://arxiv.org/abs/2312.10461}, 
}

@misc{liu2023fatformer,
      title={Forgery-aware Adaptive Transformer for Generalizable Synthetic Image Detection}, 
      author={Huan Liu and Zichang Tan and Chuangchuang Tan and Yunchao Wei and Yao Zhao and Jingdong Wang},
      year={2023},
      eprint={2312.16649},
      archivePrefix={arXiv},
      primaryClass={cs.CV},
      url={https://arxiv.org/abs/2312.16649}, 
}

@misc{wang2020cnndetect,
      title={CNN-generated images are surprisingly easy to spot... for now}, 
      author={Sheng-Yu Wang and Oliver Wang and Richard Zhang and Andrew Owens and Alexei A. Efros},
      year={2020},
      eprint={1912.11035},
      archivePrefix={arXiv},
      primaryClass={cs.CV},
      url={https://arxiv.org/abs/1912.11035}, 
}

@misc{corvi2022dmid,
      title={On the detection of synthetic images generated by diffusion models}, 
      author={Riccardo Corvi and Davide Cozzolino and Giada Zingarini and Giovanni Poggi and Koki Nagano and Luisa Verdoliva},
      year={2022},
      eprint={2211.00680},
      archivePrefix={arXiv},
      primaryClass={cs.CV},
      url={https://arxiv.org/abs/2211.00680}, 
}

@misc{frank2020freqnet,
      title={Leveraging Frequency Analysis for Deep Fake Image Recognition}, 
      author={Joel Frank and Thorsten Eisenhofer and Lea Schönherr and Asja Fischer and Dorothea Kolossa and Thorsten Holz},
      year={2020},
      eprint={2003.08685},
      archivePrefix={arXiv},
      primaryClass={cs.CV},
      url={https://arxiv.org/abs/2003.08685}, 
}

@misc{ojha2024univfd,
      title={Towards Universal Fake Image Detectors that Generalize Across Generative Models}, 
      author={Utkarsh Ojha and Yuheng Li and Yong Jae Lee},
      year={2024},
      eprint={2302.10174},
      archivePrefix={arXiv},
      primaryClass={cs.CV},
      url={https://arxiv.org/abs/2302.10174}, 
}

@misc{ju2022fusing,
      title={Fusing Global and Local Features for Generalized AI-Synthesized Image Detection}, 
      author={Yan Ju and Shan Jia and Lipeng Ke and Hongfei Xue and Koki Nagano and Siwei Lyu},
      year={2022},
      eprint={2203.13964},
      archivePrefix={arXiv},
      primaryClass={cs.CV},
      url={https://arxiv.org/abs/2203.13964}, 
}

@misc{liu2020gramnet,
      title={Global Texture Enhancement for Fake Face Detection in the Wild}, 
      author={Zhengzhe Liu and Xiaojuan Qi and Philip Torr},
      year={2020},
      eprint={2002.00133},
      archivePrefix={arXiv},
      primaryClass={cs.CV},
      url={https://arxiv.org/abs/2002.00133}, 
}

@misc{koutlis2024rine,
      title={Leveraging Representations from Intermediate Encoder-blocks for Synthetic Image Detection}, 
      author={Christos Koutlis and Symeon Papadopoulos},
      year={2024},
      eprint={2402.19091},
      archivePrefix={arXiv},
      primaryClass={cs.CV},
      url={https://arxiv.org/abs/2402.19091}, 
}

@misc{sha2023defake,
      title={DE-FAKE: Detection and Attribution of Fake Images Generated by Text-to-Image Generation Models}, 
      author={Zeyang Sha and Zheng Li and Ning Yu and Yang Zhang},
      year={2023},
      eprint={2210.06998},
      archivePrefix={arXiv},
      primaryClass={cs.CR},
      url={https://arxiv.org/abs/2210.06998}, 
}

@misc{zhong2024patchcraft,
      title={PatchCraft: Exploring Texture Patch for Efficient AI-generated Image Detection}, 
      author={Nan Zhong and Yiran Xu and Sheng Li and Zhenxing Qian and Xinpeng Zhang},
      year={2024},
      eprint={2311.12397},
      archivePrefix={arXiv},
      primaryClass={cs.CV},
      url={https://arxiv.org/abs/2311.12397}, 
}

@INPROCEEDINGS{pellegrini2025aigenbench,
  author={Pellegrini, Lorenzo and Cozzolino, Davide and Pandolfini, Serafino and Maltoni, Davide and Ferrara, Matteo and Verdoliva, Luisa and Prati, Marco and Ramilli, Marco},
  booktitle={2025 International Joint Conference on Neural Networks (IJCNN)}, 
  title={AI-GenBench: A New Ongoing Benchmark for AI-Generated Image Detection}, 
  year={2025},
  volume={},
  number={},
  pages={1-9},
  doi={10.1109/IJCNN64981.2025.11228377}}

@misc{cozzolino2018noiseprint,
      title={Noiseprint: a CNN-based camera model fingerprint}, 
      author={Davide Cozzolino and Luisa Verdoliva},
      year={2018},
      eprint={1808.08396},
      archivePrefix={arXiv},
      primaryClass={cs.CV},
      url={https://arxiv.org/abs/1808.08396}, 
}

@INPROCEEDINGS{frick2024diffseg,
  author={Frick, Raphael Antonius and Steinebach, Martin},
  booktitle={2024 IEEE/CVF Conference on Computer Vision and Pattern Recognition Workshops (CVPRW)}, 
  title={DiffSeg: Towards Detecting Diffusion-Based Inpainting Attacks Using Multi-Feature Segmentation}, 
  year={2024},
  volume={},
  number={},
  pages={3802-3808},
  doi={10.1109/CVPRW63382.2024.00384}}

@misc{xu2025fakeshield,
      title={FakeShield: Explainable Image Forgery Detection and Localization via Multi-modal Large Language Models}, 
      author={Zhipei Xu and Xuanyu Zhang and Runyi Li and Zecheng Tang and Qing Huang and Jian Zhang},
      year={2025},
      eprint={2410.02761},
      archivePrefix={arXiv},
      primaryClass={cs.CV},
      url={https://arxiv.org/abs/2410.02761}, 
}

@misc{li2022mat,
      title={MAT: Mask-Aware Transformer for Large Hole Image Inpainting}, 
      author={Wenbo Li and Zhe Lin and Kun Zhou and Lu Qi and Yi Wang and Jiaya Jia},
      year={2022},
      eprint={2203.15270},
      archivePrefix={arXiv},
      primaryClass={cs.CV},
      url={https://arxiv.org/abs/2203.15270}, 
}

@misc{ju2024brushnet,
      title={BrushNet: A Plug-and-Play Image Inpainting Model with Decomposed Dual-Branch Diffusion}, 
      author={Xuan Ju and Xian Liu and Xintao Wang and Yuxuan Bian and Ying Shan and Qiang Xu},
      year={2024},
      eprint={2403.06976},
      archivePrefix={arXiv},
      primaryClass={cs.CV},
      url={https://arxiv.org/abs/2403.06976}, 
}

@InProceedings{Rombach_2022_CVPR,
    author    = {Rombach, Robin and Blattmann, Andreas and Lorenz, Dominik and Esser, Patrick and Ommer, Bj\"orn},
    title     = {High-Resolution Image Synthesis With Latent Diffusion Models},
    booktitle = {Proceedings of the IEEE/CVF Conference on Computer Vision and Pattern Recognition (CVPR)},
    month     = {June},
    year      = {2022},
    pages     = {10684-10695}
}

@misc{AdobeFirefly2025,
  title        = {Adobe Firefly},
  howpublished = {\url{https://www.adobe.com/it/products/firefly/landpa.html}},
  note         = {Access: 2025-12-03}
}

@misc{zhuang2024powerpaint,
      title={A Task is Worth One Word: Learning with Task Prompts for High-Quality Versatile Image Inpainting}, 
      author={Junhao Zhuang and Yanhong Zeng and Wenran Liu and Chun Yuan and Kai Chen},
      year={2024},
      eprint={2312.03594},
      archivePrefix={arXiv},
      primaryClass={cs.CV},
      url={https://arxiv.org/abs/2312.03594}, 
}

@misc{BlackForestFlux2024,
  author       = {{Black Forest Labs}},
  title        = {Flux},
  howpublished = {\url{https://github.com/black-forest-labs/flux}},
  year         = {2024},
  note         = {Access: 2025-12-03}
}

@misc{TGIF2_2024,
  author       = {{IDLabMedia}},
  title        = {TGIF: Text‑Guided Inpainting Forgery Dataset},
  howpublished = {\url{https://github.com/IDLabMedia/tgif-dataset}},
  year         = {2024},
  note         = {Access: 2025-12-03}
}

@ARTICLE{wu2021inaintinglocalizationdataset,
author={Wu, Haiwei and Zhou, Jiantao},
journal={IEEE Transactions on Circuits and Systems for Video Technology},
title={IID-Net: Image Inpainting Detection Network via Neural Architecture Search and Attention},
year={2021},
volume={},
number={},
pages={1-1},
doi={10.1109/TCSVT.2021.3075039}
}

@misc{roesch2024inpaintcoco,
      title={Enhancing Conceptual Understanding in Multimodal Contrastive Learning through Hard Negative Samples}, 
      author={Philipp J. Rösch and Norbert Oswald and Michaela Geierhos and Jindřich Libovický},
      year={2024},
      eprint={2403.02875},
      archivePrefix={arXiv},
      primaryClass={cs.CV}
}

@misc{aboutalebi2024deepfakeart,
      title={DeepfakeArt Challenge: A Benchmark Dataset for Generative AI Art Forgery and Data Poisoning Detection}, 
      author={Hossein Aboutalebi and Dayou Mao and Rongqi Fan and Carol Xu and Chris He and Alexander Wong},
      year={2024},
      eprint={2306.01272},
      archivePrefix={arXiv},
      primaryClass={cs.CV},
      url={https://arxiv.org/abs/2306.01272}, 
}

@inproceedings{bertazzini2024beyondthebush,
  title={Beyond the Brush: Fully-automated Crafting of Realistic Inpainted Images},
  author={Bertazzini, Giulia and Albisani, Chiara and Baracchi, Daniele and Shullani, Dasara and Piva, Alessandro},
  booktitle={2024 IEEE International Workshop on Information Forensics and Security (WIFS)},
  pages={1--6},
  year={2024},
  organization={IEEE},
  doi={10.1109/WIFS61860.2024.10810722}
}

@inproceedings{GC,
title={Free-form image inpainting with gated convolution},
author={J. H. Yu and Z. Lin and J. M. Yang and X. H. Shen and X. Lu and T. S. Huang},
booktitle={Proc. IEEE Int. Conf. Comput. Vis.},
pages={4471--4480},
year={2019}
}

@inproceedings{CA,
title={Generative image inpainting with contextual attention},
author={J. H. Yu and Z. Lin and J. M. Yang and X. H. Shen and X. Lu and T. S. Huang},
booktitle={Proc. IEEE Conf. Comput. Vis. Pattern Recogn.},
pages={5505--5514},
year={2018}
}

@inproceedings{SH,
title={Shift-net: image inpainting via deep feature rearrangement},
author={Z. Y. Yan and X. M. Li and M. Li and W. M. Zuo and S. G. Shan},
booktitle={Proc. Eur. Conf. Comput. Vis.},
pages={1--17},
year={2018}
}

@inproceedings{EC,
title={EdgeConnect: generative image inpainting with adversarial edge learning},
author={K. Nazeri and E. Ng and T. Joseph and F. Z. Qureshi and M. Ebrahimi},
booktitle={Proc. IEEE Int. Conf. Comput. Vis. Workshop},
year={2019},
}

@inproceedings{RN,
title={Region normalization for image inpainting},
author={T. Yu and Z. Guo and X. Jin and S. Wu and Z. Chen and W. Li and Z. Zhang and S. Liu},
booktitle={Proc. AAAI Conf. Arti. Intell.},
pages={12733--12740},
year={2020}
}

@article{TE,
title={An image inpainting technique based on the fast marching method},
author={A. Telea},
journal={J. of Graph. Tools},
volume={9},
number={1},
pages={23--34},
year={2004},
publisher={Taylor \& Francis}
}

@inproceedings{NS,
title={Navier-stokes, fluid dynamics, and image and video inpainting},
author={M. Bertalmio and A. L. Bertozzi and G. Sapiro},
booktitle={Proc. IEEE Conf. Comput. Vis. Pattern Recogn.},
pages={355--362},
year={2001}
}

@article{PM,
title={High-quality real-time video inpaintingwith PixMix},
author={J. Herling and W. Broll},
journal={IEEE Trans. Vis. Comput. Graph.},
volume={20},
number={6},
pages={866--879},
year={2014},
publisher={IEEE}
}

@article{SG,
title={Image completion using planar structure guidance},
author={J. Huang and S. Kang and N. Ahuja and J. Kopf},
journal={ACM Trans. Graph.},
volume={33},
number={4},
pages={1--10},
year={2014},
publisher={ACM}
}

@article{LR,
title={Patch-based image inpainting via two-stage low rank approximation},
author={Q. Guo and S. Gao and X. Zhang and Y. Yin and C. Zhang},
journal={IEEE Trans. Vis. Comput. Graph.},
volume={24},
number={6},
pages={2023--2036},
year={2017},
publisher={IEEE}
}

@misc{lin2015microsoftcococommonobjects,
      title={Microsoft COCO: Common Objects in Context}, 
      author={Tsung-Yi Lin and Michael Maire and Serge Belongie and Lubomir Bourdev and Ross Girshick and James Hays and Pietro Perona and Deva Ramanan and C. Lawrence Zitnick and Piotr Dollár},
      year={2015},
      eprint={1405.0312},
      archivePrefix={arXiv},
      primaryClass={cs.CV},
      url={https://arxiv.org/abs/1405.0312}, 
}

@INPROCEEDINGS{imagenet2009,
  author={Deng, Jia and Dong, Wei and Socher, Richard and Li, Li-Jia and Kai Li and Li Fei-Fei},
  booktitle={2009 IEEE Conference on Computer Vision and Pattern Recognition}, 
  title={ImageNet: A large-scale hierarchical image database}, 
  year={2009},
  volume={},
  number={},
  pages={248-255},
  doi={10.1109/CVPR.2009.5206848}}

@ARTICLE{places2018,
  author={Zhou, Bolei and Lapedriza, Agata and Khosla, Aditya and Oliva, Aude and Torralba, Antonio},
  journal={IEEE Transactions on Pattern Analysis and Machine Intelligence}, 
  title={Places: A 10 Million Image Database for Scene Recognition}, 
  year={2018},
  volume={40},
  number={6},
  pages={1452-1464},
  doi={10.1109/TPAMI.2017.2723009}}

@misc{radford2021resnet50,
      title={Learning Transferable Visual Models From Natural Language Supervision}, 
      author={Alec Radford and Jong Wook Kim and Chris Hallacy and Aditya Ramesh and Gabriel Goh and Sandhini Agarwal and Girish Sastry and Amanda Askell and Pamela Mishkin and Jack Clark and Gretchen Krueger and Ilya Sutskever},
      year={2021},
      eprint={2103.00020},
      archivePrefix={arXiv},
      primaryClass={cs.CV},
      url={https://arxiv.org/abs/2103.00020}, 
}

@misc{oquab2024dinov2,
      title={DINOv2: Learning Robust Visual Features without Supervision}, 
      author={Maxime Oquab and Timothée Darcet and Théo Moutakanni and Huy Vo and Marc Szafraniec and Vasil Khalidov and Pierre Fernandez and Daniel Haziza and Francisco Massa and Alaaeldin El-Nouby and Mahmoud Assran and Nicolas Ballas and Wojciech Galuba and Russell Howes and Po-Yao Huang and Shang-Wen Li and Ishan Misra and Michael Rabbat and Vasu Sharma and Gabriel Synnaeve and Hu Xu and Hervé Jegou and Julien Mairal and Patrick Labatut and Armand Joulin and Piotr Bojanowski},
      year={2024},
      eprint={2304.07193},
      archivePrefix={arXiv},
      primaryClass={cs.CV},
      url={https://arxiv.org/abs/2304.07193}, 
}

@misc{siméoni2025dinov3,
      title={DINOv3}, 
      author={Oriane Siméoni and Huy V. Vo and Maximilian Seitzer and Federico Baldassarre and Maxime Oquab and Cijo Jose and Vasil Khalidov and Marc Szafraniec and Seungeun Yi and Michaël Ramamonjisoa and Francisco Massa and Daniel Haziza and Luca Wehrstedt and Jianyuan Wang and Timothée Darcet and Théo Moutakanni and Leonel Sentana and Claire Roberts and Andrea Vedaldi and Jamie Tolan and John Brandt and Camille Couprie and Julien Mairal and Hervé Jégou and Patrick Labatut and Piotr Bojanowski},
      year={2025},
      eprint={2508.10104},
      archivePrefix={arXiv},
      primaryClass={cs.CV},
      url={https://arxiv.org/abs/2508.10104}, 
}

\end{document}